\documentclass[acmtog, authorversion, screen]{acmart}
\acmJournal{TOG}
\acmYear{2024} \acmVolume{0} \acmNumber{0} \acmArticle{0} \acmMonth{1} \acmPrice{}\acmDOI{}
\setcopyright{none}

\usepackage{booktabs} % For formal tables

\usepackage{graphicx}
\usepackage{hyperref}
\AtEndPreamble{
    \usepackage[capitalize]{cleveref}
    \crefname{section}{Sec.}{Secs.}
    \Crefname{section}{Section}{Sections}
    \Crefname{table}{Table}{Tables}
    \crefname{table}{Tab.}{Tabs.}
}

\newcommand{\eg}{\textit{e}.\textit{g}. }

\title{AvatarMMC: 3D Head Avatar Generation and Editing with Multi-Modal Conditioning}

\author{Wamiq Reyaz Para}
\affiliation{
        {\institution{KAUST}
           \country{Saudi Arabia}
            }
        }
\author{Abdelrahman Eldesokey}
\affiliation{
        {\institution{KAUST}
           \country{Saudi Arabia}
            }
        }
\author{Zhenyu Li}
\affiliation{
        {\institution{KAUST}
           \country{Saudi Arabia}
            }
        }
\author{Pradyumna Reddy}
\affiliation{
        {\institution{University College London}
           \country{UK}
            }
        }

\author{Jiankang Deng}
\affiliation{
        {\institution{Imperial College London}
           \country{UK}
            }
        }

\author{Peter Wonka}
\affiliation{
        {\institution{KAUST}
           \country{Saudi Arabia}
            }
        }

\begin{document}

\begin{teaserfigure}
    \centering
     \includegraphics[width=\linewidth]{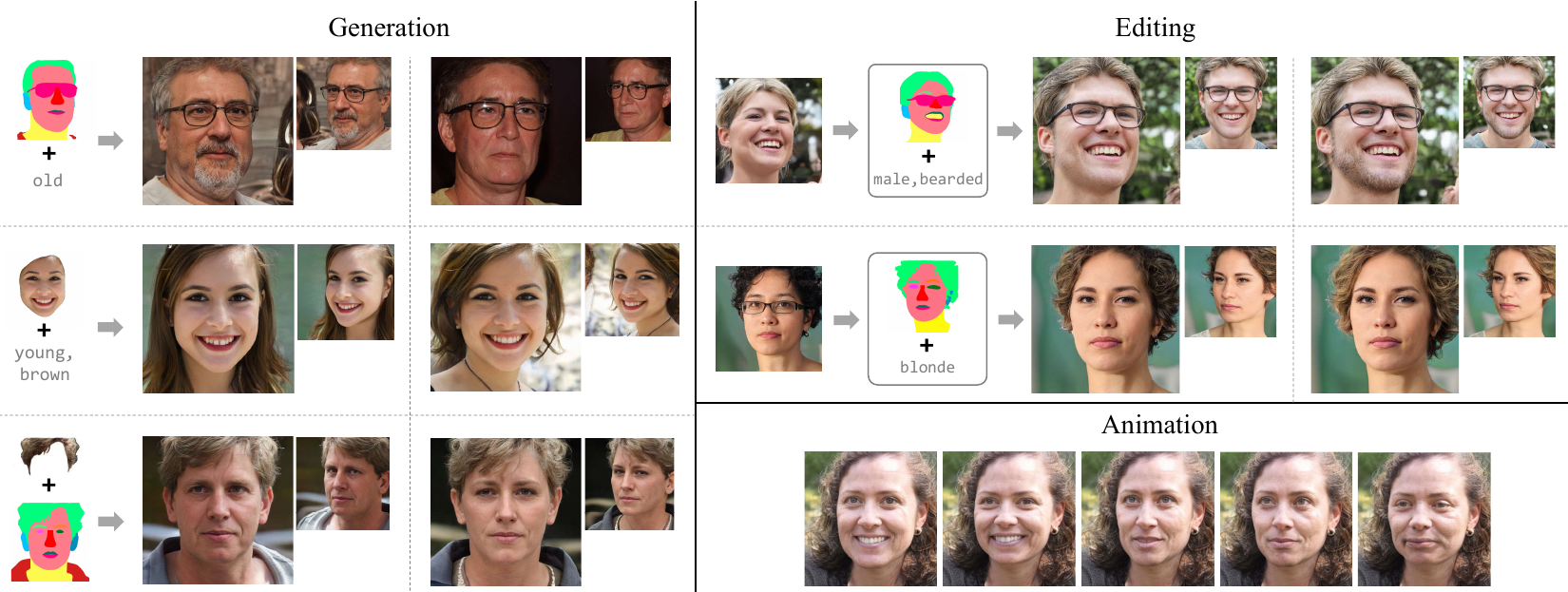}
    \captionof{figure}{Our approach can generate high-quality 3D animatable head avatars given a combination of \emph{multi-modal control signals} such as attributes, face segmentation maps, and RGB information. It can also perform multi-modal conditional editing with high-fidelity.}
    \label{fig:teaser}
\end{teaserfigure}

\begin{abstract}

We introduce an approach for 3D head avatar generation and editing with \emph{multi-modal conditioning} based on a 3D Generative Adversarial Network (GAN) and a Latent Diffusion Model (LDM).
3D GANs can generate high-quality head avatars given a single or no condition.
However, it is challenging to generate samples that adhere to multiple conditions of different modalities.
On the other hand, LDMs excel at learning complex conditional distributions.
To this end, we propose to exploit the conditioning capabilities of LDMs to enable multi-modal control over the latent space of a pre-trained 3D GAN.
Our method can generate and edit 3D head avatars given a mixture of control signals such as RGB input, segmentation masks, and global attributes.
This provides better control over the generation and editing of synthetic avatars both globally and locally.
Experiments show that our proposed approach outperforms a solely GAN-based approach both qualitatively and quantitatively on generation and editing tasks.
To the best of our knowledge, our approach is the first to introduce multi-modal conditioning to 3D avatar generation and editing.
% Experimental results show that we outperform the baseline in editing and generation tasks.
%
% Combining the generative modeling capability of the latent diffusion model with a 3D-GAN allows us to use the high-quality generation of a GAN model while enabling multi-modal conditioning and editing by modeling latent vectors of a 3D GAN model trained exclusively on 2D images.  
% The conditional generation and editing versatility showcased by our method marks a significant improvement in controllable facial avatar synthesis. 
\end{abstract}
\maketitle

\section{Introduction}
\label{sec:intro}

% The best generative models to generate 3D head avatars are 3D GANs \cite a bunch.
% 11:43
% We would like to tackle the problem of introducing conditioning information for better control of the generation, editing, and animation.
% 11:44
% The requirements are that the conditioning can be multi-modal, allows for multiple types of conditioning jointly, and all conditioning information is optional so that any subset of conditioning information can be used without retraining the model.
% 11:46
% Then I would focus on reviewing different approaches to do that. Say that many approaches do not fulfill all the criteria and cite some examples. Then I would review the apple paper in more detail and list all our advantages, as well as explain the idea of optimization as a baseline to compare to.
% 11:46
% Then list the main ideas of our approach
% 11:46
% and the contributions

Generating 3D head avatars is a pivotal task in movie production, social media, and visual effects industries with many downstream applications, \eg, creating synthetic characters, editing existing ones, and animating them.
Therefore, it is deemed favorable to develop approaches that are capable of generating high-quality avatars.
At the same time, these approaches should possess a high degree of controllability over the generated avatars to fulfill the needs of creators.
3D Generative Adversarial Network (GAN) \cite{chan2022efficient,sun2023next3d,zhang2022multi} have demonstrated outstanding capabilities in generating highly-realistic avatars.
Furthermore, several GAN-based approaches \cite{gal2022stylegan,lin2023sketchfacenerf,gu2023learning} were proposed to condition the generation process on a given signal, \eg, text, segmentation masks, or sketches.
However, all these approaches are capable of incorporating only a \emph{single} condition limiting their controllability.

A natural solution to this problem is incorporating multiple conditions into an optimization-based pipeline.
Nonetheless, we found that this approach performs badly and introduces the complexity of defining the optimization objective.
A notable work is Control3Diff \cite{gu2023learning}, which proposed to combine 3D GANs with a diffusion model \cite{dhariwal2021diffusion} to generate 3D avatars either unconditionally or conditionally.
However, their conditional variation can only accept a single \emph{uni-modal} condition, \eg, text, sketch, or a segmentation map, limiting their controllability.
It also requires training both the 3D GAN and the 2D diffusion model, which is computationally expensive.
Moreover, the generated avatars are not animatable due to generating a tri-plane directly without accounting for a mesh as in Next3D \cite{sun2023next3d}.

In this paper, we tackle these shortcomings, and we propose an approach for animatable 3D head avatar generation and editing with \emph{multi-modal conditioning}.
We use a pre-trained 3D-GAN \cite{sun2023next3d} for avatar generation, and we design a 1D Latent Diffusion Model (LDM) \cite{stable_diffusion} to incorporate multi-modal conditioning over the latent space of the GAN.
This allows specifying a set of control signals, such as partial RGB information, face segmentation masks, and global attributes, to control and edit the generated avatar. 
Our proposed approach does not require training the GAN, and it is fast to sample from.
Experiments show that our approach outperforms an optimization-based GAN baseline both on avatar generation and editing tasks.

\noindent In summary, we have the following contributions:
\begin{enumerate}
    \item We propose a framework for 3D avatar generation and editing with \emph{multi-modal conditioning}.
    \item Our approach employs the latent space of a 3D GAN for avatar representation and an LDM for imposing multi-modal control over this latent space. 
    \item Experiments demonstrate that our approach outperforms a purely GAN-based approach on avatar generation and editing.
\end{enumerate}

\begin{figure*}[!t]
    \centering
     \includegraphics[width=\linewidth, height=2in]{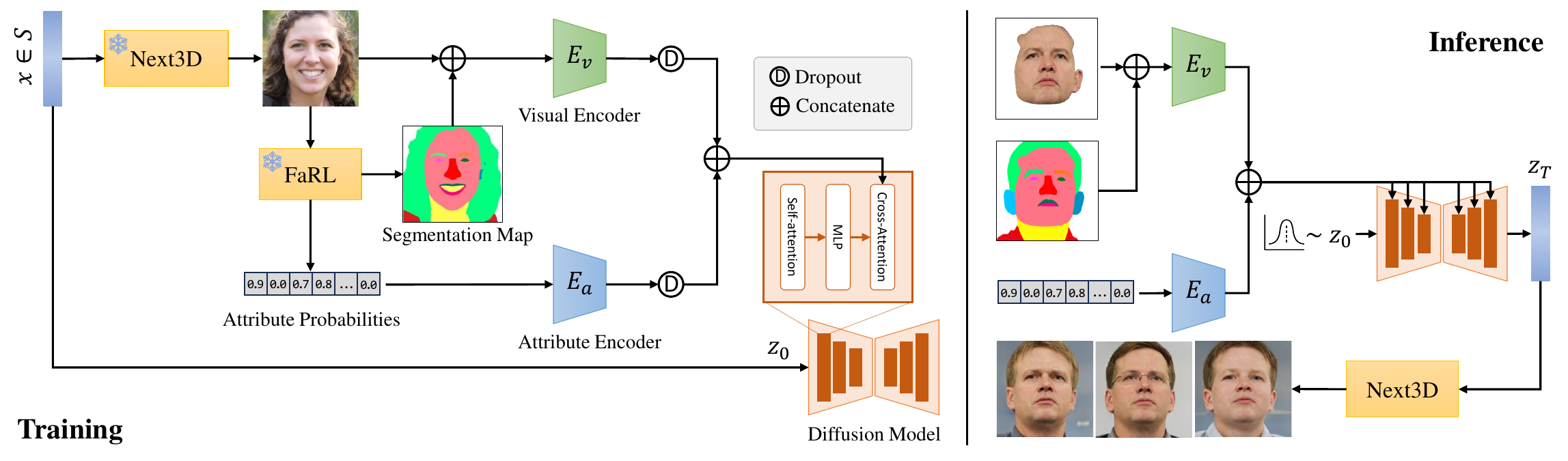}
    \captionof{figure}{ \textbf{Overview of the method.} We use the StyleSpace $\mathcal{S}$ of the pre-trained 3D GAN, Next3D \cite{sun2023next3d}, as our avatar space, and we train a diffusion model to impose \emph{multi-conditional control} over this space. During inference, any combination of the conditions can be used to control the sampling process. Our method can produce multiple diverse results that satisfy the provided conditions.}
    \label{fig:pipeline}
\end{figure*}

% (\textbf{Left}): We use the Style-space  of a 3D-GAN as our 3D prior and learn a Composable Latent Diffusion Model over the space of this GAN. Our model can control the generation process at multiple levels of control. From coarsest to finest they are: attribute level, segmentation level and image level. (\textbf{Middle}): These conditions are encoded jointly into a common condition space. Our proposed architecture relies on training the conditional diffusion model where the conditions are applied to the denoising process with an attention mechanism. (\textbf{Right}): With our sampling scheme (Section~\ref{sec:sampling}), inference can be performed with any subset of the control modalities, giving the user the desired control over the generation process by injecting the generated style-codes back into the 3D-GAN. 

\section{Related Work}
\label{sec:formatting}
\subsection{3D Generative Adversarial Networks (GANs)}
% GANs~\cite{goodfellow2014generative} have been successful in generating photorealistic images in the 2D domain. 
% Expanding on this success, there have been numerous works that have extended image synthesis to the 3D domain. 
Early 3D GANs were predominantly voxel-based~\cite{gadelha20173d, henzler2019escaping, nguyen2019hologan, nguyen2020blockgan} and used 3D convolutional neural networks as generators. 
However, these methods had significant computational demands, which limited their ability to generate high-resolution images. 
Recent methods used fully implicit networks~\cite{cai2022pix2nerf, chan2022efficient, chen2022gdna, chen2022sem2nerf, deng2022gram, xu2021generative, schwarz2020graf, skorokhodov2022epigraf, zhou2021cips}, sparse voxel grids~\cite{schwarz2022voxgraf}, and multiple planes~\cite{kplanes_2023, zhao2022generative}, to make scene representations efficient by combining low-resolution feature volumes and 2D super-resolution~\cite{chan2022efficient, gu2021stylenerf, niemeyer2021giraffe, xue2022giraffe, or2022stylesdf, xu20223d,  zhang20223d, zhang2022multi}. 
This technique successfully improved unconditional 3D generation performance at a reasonable computational cost. 
In this paper, we extend an unconditional 3D GAN to perform multi-modal conditional generation efficiently.

%%%%%%%%%%%%%%%%%%%%%%%%%%%%%%%%%%%%%%%%%%%%%%%%%%%%%%%%%%%%%%%%%%%%%%%%%%

\subsection{Conditional Generative Models}
Conditional generation~\cite{ding2021cogview, esser2021imagebart, gafni2022make, koh2021text, huang2023collaborative, liu2023more, ramesh2021zero, ramesh2022hierarchical, reed2016generative, reed2016learning, saharia2022photorealistic, wang2022pretraining, xia2021tedigan, xia2021towards, xu2018attngan}, and editing~\cite{couairon2022diffedit, lin2023sketchfacenerf, lee2020maskgan, patashnik2021styleclip, shen2020interfacegan, shen2020interpreting, xia2021tedigan} are actively researched areas.
Most work focused on conditioning generative models using individual modalities like text~\cite{gal2022stylegan, jiang2022text2human, patashnik2021styleclip, xia2021tedigan,nie2021controllable}, segmentation masks~\cite{lee2020maskgan, li2020manigan, park2019semantic}, audio inputs~\cite{song2018talking}, and sketches~\cite{lin2023sketchfacenerf}.

Several approaches were proposed to enable multi-modal control in diffusion models \cite{zhang2023adding,mou2023t2i} by exploiting their ability to learn conditional distributions.
Control3Diff \cite{gu2023learning} leveraged this to train a controllable diffusion model for generating 3D avatars conditioned on different modalities.
However, it cannot simultaneously handle multiple conditions, like identity, textual attributes, and segmentation masks, and it requires re-training the GAN model.
It also trains the diffusion model on the tri-plane representation with a high dimensionality of $256 \times 256 \times 32$, which is computationally expensive to train and sample from.
Instead, we propose a lighter 1D Latent Diffusion Model (LDM) to impose control over the latent space of a pre-trained 3D GAN that requires no re-training of the GAN.
Our diffusion model operates on the latent space of the 3D GAN with a dimensionality of $512 \times 73$, which is 50 times lower than this of Control3Diff.
Furthermore, our proposed model enables \emph{multi-modal conditioning} for 3D avatar generation and editing, and it combines the quality of 3D GANs with the controllability of diffusion models.

%%%%%%%%%%%%%%%%%%%%%%%%%%%%%%%%%%%%%%%%%%%%%%%%%%%%%%%%%%%%%%%%%%%%%%%%%%

\section{Preliminaries}
\subsection{3D GANs for Head Avatars}
\label{sec:3dgan}

We give a brief overview of the 3D GAN, EG3D~\cite{chan2022efficient} and its extension Next3D~\cite{sun2023next3d}.
EG3D uses a standard StyleGAN2~\cite{Karras2019AnalyzingAI} generator to generate features for a so-called \emph{tri-plane}. 
This tri-plane together with an MLP defines a neural radiance field~\cite{mildenhall2020nerf} which is rendered to generate an image.
% Through proper regularization and clever discriminator design, the features are view-consistent. 
% As a result, the generator is able to learn a 3D, multi-view consistent representation using only posed 2D images. 
The generated images are high quality but the generation process is unconditional. 
Furthermore, once a 3D representation has been generated, animating it is a formidable challenge due to the lack of intuitive control, \eg based on a mesh.

Next3D adds animation to the base EG3D model by ensuring the tri-plane features depend on a FLAME model \cite{FLAME:SiggraphAsia2017}. 
In particular, during training, Next3D samples a FLAME mesh with a random identity and expression. 
This mesh is rasterized onto the tri-plane, and the original triplane features are replaced by a texture lookup of the rasterized image. 
The dual-discriminator of EG3D is modified so that instead of only generating a score for whether the generated image is real, it also generates a score for whether the generated image matches the identity and expression of the sampled FLAME mesh. 
We built on Next3D in our work. 

\noindent \textbf{Editing Head Avatars in GANs}: GANs are not likelihood-based models, and consequently, we do not have access to a prior over its latents. 
Editing images with GANs has therefore relied on inversion~\cite{Abdal2019Image2StyleGANHT, sg++, roich2021pivotal} of an image into the latent-space, and then manipulating the latents~\cite{Wu2020StyleSpaceAD} to achieve edits on the \textit{embedded} images. 
There are many targets for the latent space, but we use the StyleSpace $\mathcal{S}$~\cite{Wu2020StyleSpaceAD} as our latent space so that $ \{\mathbf{x} \in \mathcal{R}^{73 \times 512} | \mathbf{x} \in \mathcal{S}\}$, where $\mathbf{x}$ is a data sample.
We consider this to be a 1D sequence of the 73 style-codes each of which is 512-dimensional, and modulate each of the StyleGAN convolutions in the generator. 
Style-codes that are not 512-dimensional are simply zero-padded.

%%%%%%%%%%%%%%%%%%%%%%%%%%%%%%%%%%%%%%%%%%%%%%%%%%%%%%%%%%%%%%%%%%%%%%%%%%

\subsection{Latent Diffusion Models (LDMs)}
\label{sec:diffusion}
Denoising-diffusion models~\cite{SohlDickstein2015DeepUL} are a class of generative models that generate data by an iterative Markovian denoising process, starting from isotropic Gaussian noise. 
Given data $\mathbf{x}_0$, a \textit{forward process} adds Gaussian noise to the data progressively over timesteps $t$ producing noisy data $\mathbf{x}_t$ with $t \in [1, T]$. 
\begin{equation}
    q(\mathbf{x}_t|\mathbf{x}_{t-1}) = \mathcal{N}(\mathbf{x}_t; \alpha_t \mathbf{x}_{t-1}, \sigma^2_{t}\mathbf{I})
\end{equation}
Here, $T$ is the maximum number of steps, $\alpha_t = \sqrt{1 - \beta_t}$ and $\sigma^2_t = \beta_t$ where $\beta_1, \ldots \beta_T$ are generated according to a fixed variance schedule. 
This schedule ensures that $\mathbf{x}_T$, for a sufficiently large $T$, is distributed as an isotropic Gaussian. 
The diffusion model $\epsilon_\theta$ with parameters $\theta$, which is typically a UNet, learns the \textit{reverse-process} which attempts to denoise every step by predicting $\epsilon_t$, the noise added at each step $t$ during the forward process. 
The loss is defined as the weighted expectation of minimizing the difference between the added noise and the predicted noise at each timestep $t$:
\begin{equation}
    \mathcal{L}_{diff} = \mathbb{E}_{\mathbf{x}_t \sim q(\mathbf{x}_t|\mathbf{x}_0) \, t \sim \mathcal{U}\{1\ldots T\}} \left[|| \epsilon_t - \epsilon_\theta(\mathbf{x}_t, t)||^2 \right]
    \label{eq:l_diff}
\end{equation}
This formula describes an unconditional diffusion model, but it can be adapted to incorporate some arbitrary conditions $\mathbf{c}$.
In addition, we can work in a compressed space~\cite{Rombach2021HighResolutionIS} $z_0$ for efficiency such that $\mathbf{x}_0 = \mathcal{D}(\mathbf{z}_0)$ and $\mathbf{z}_0 = \mathcal{E}(\mathbf{x}_0)$, where $\mathcal{E}$ and $\mathcal{D}$ are some pre-trained encoder and decoder, respectively.
This leads to the following formulation:
\begin{equation}
    \mathcal{L}_{LDM} = \mathbb{E}_{\mathbf{z}_t \sim q(\mathbf{z}_t|\mathbf{z}_0) \, t \sim \mathcal{U}\{1\ldots T\}} \left[|| \epsilon_t - \epsilon_\theta(\mathbf{z}_t, \mathbf{c}, t)||^2 \right]
    \label{eq:l_diff2}
\end{equation}
Sampling is performed with ancestral sampling, sequentially generating $\mathbf{z}_T, \mathbf{z}_{T-1}, \ldots, \mathbf{z}_0$, where $\mathbf{z}_T \sim \mathcal{N}(\mathbf{0}, \mathbf{I})$.

\noindent \textbf{Classifier-Free Guidance}: For sampling, we use \textit{Classifier-free guidance} (CFG)~\cite{Ho2022ClassifierFreeDG}. 
This technique has been successfully used as a sampling method, particularly in text-conditioned generation~\cite{Ramesh2022HierarchicalTI,saharia2022photorealistic}. 
CFG requires minor tweaks to the training and inference schemes implied by Eq. \ref{eq:l_diff2}. 
At training time, instead of training a purely conditional model, the diffusion model is also trained unconditionally by replacing the condition $\mathbf{c}$ with a null condition $\phi$ randomly.
For inference, CFG is implemented by changing the predicted noise at every denoising step. 
In order to generate data that adheres to condition $\mathbf{c}$, we use a weight $\omega$ and modify the estimated noise 
\begin{equation}
    \tilde{\epsilon}_\theta(\mathbf{z}_t, \mathbf{c}, t) = (1 - \omega){\epsilon}_\theta(\mathbf{z}_t, \phi, t) + \omega {\epsilon}_\theta(\mathbf{z}_t, \mathbf{c}, t)
    \label{eq:cfg}
\end{equation}
Higher values of $\omega$ tend to correlate the output more with the input condition $\mathbf{c}$ but at the cost of the quality of the generated samples.

%% TODO:

% 4 V100 (1 node) or a single A100
% Cite nie2021controllables

\section{Method}
\label{sec:method}
We base our approach on a pre-trained Next3D \cite{sun2023next3d} as described in \Cref{sec:3dgan}, and we seek to achieve multi-modal control over the sampling process.  
In this work, we consider three conditional modalities, partial RGB input and face segmentation maps for local control and attributes for global control.
This allows control of the generative process at three different levels: \emph{coarse}, where we only specify the attributes (\eg, blonde, fair, young), \emph{mid}, where the shape is specified by face segmentation masks, and \emph{fine} where the RGB appearance is specified to encode specific details of a particular identity. 
However, our approach can be adapted to incorporate other conditioning modalities. 
Note that our approach can compose these conditioning signals, \textit{i.e.}, one can specify some or none of these modalities.
The overview of the method is shown in Figure \ref{fig:pipeline}.

% Our goal is to distill the 3D knowledge of the pre-trained 3D-GAN, the attribute classifier, and the segmentation model into a diffusion model over the space of 3D-GAN latents. 

% Simultaneously, a Visual Encoder is trained to encode the \emph{unposed} RGB and segmentation corresponding to the latent so that our diffusion model can be controlled by the final appearance of the model from a particular view. 
% The Attribute Encoder performs the task of encoding the attributes. 
% Then, armed with Classifer-Free Guidance (CFG), we are able to control the generated 3D face. 

% We further train our diffusion model to be able to compose these controlling inputs - one can specify the shape and the attributes only, or one can specify the attributes and some of the color values. 
% The inputs themselves are masked, allowing for only partial specification - only the blondness and not the skin color for attributes or only the hair color but not the skin color when specifying the color values. 
% We train only on synthetically generated images and their corresponding attributes.

%%%%%%%%%%%%%%%%%%%%%%%%%%%%%%%%%%%%%%%%%%%%%%%%%%%%%%%%%%%%%%

\subsection{Embedding Conditional Modalities}
The first step to multi-modal conditioning is embedding different modalities into a common space.
We achieve this by employing a per-modality encoder, which untangles the problem and allows dynamically incorporating different modalities.

%%%%%%%%%%%%%%%%%%%%%%%%%%%

\subsubsection{Attribute Encoder}
We use the FaRL attribute predictor \cite{zheng2022farl} to extract a list of attributes for all images in the datasets that we use.
Each attribute is represented as a probability vector over all possible attributes in the dataset with dimensionality $1 \times n$, where $n$ is the number of possible attributes.
We design an encoder $\mathbf{E}_\textit{a}$ to embed the attributes and output an attribute embedding $\mathbf{c}_\textit{a}$.
For each of the $n=21$ attributes, we encode each attribute with sinusoidal-encoding with 256 levels to create a $n \times 512$ dimensional tensor to which a learnable position embedding is added. 
This is further processed with a 5-layer MLP with ReLU activations. 

%%%%%%%%%%%%%%%%%%%%%%%%%%%

\subsubsection{Visual Encoder}
For encoding the segmentation maps and the RGB input modalities, we found that our model does not benefit from having two separate encoders.
Therefore, we opt for a single visual encoder for the segmentation maps and the RGB input combined for efficiency.
We denote this encoder as $\mathbf{E}_\textit{v}$, and it generates the visual condition $\mathbf{c}_\textit{v}$.
We use a DeeplabV3+~\cite{deeplabv3plus2018} architecture pre-trained on the MS-COCO~\cite{mscoco} dataset.
The first convolution layer is modified to accept a 4-channel input, the 3-channel RGB image $\mathcal{I}$ concatenated with the single channel segmentation mask $\mathcal{M}$.
We also add a 3-layer MLP with a ReLU activation immediately after the ASPP module to extract the final features $\mathbf{c}_\textit{v}$ and project them to the same dimensionality as the attribute embedding  $\mathbf{c}_\textit{a}$.
% The pretrained model already produces high quality and high resolution visual features, thus reducing the training time by providing the diffusion model with high quality from the start of training. 

%%%%%%%%%%%%%%%%%%%%%%%%%%%%%%%%%%%%%%%%%%%%%%%%%%%%%%%%%%%%%%

\subsection{Multi-Modal Conditioning} 
Latent Diffusion Models (LDMs) \cite{stable_diffusion} have demonstrated remarkable success in synthesizing images conditioned on several modalities, such as text, sketches, and depth maps.
Inspired by this success, we adopt a diffusion-based approach for enabling multi-modal conditioning over the latent space of Next3D GAN for highly controllable avatar generation and editing.
More specifically, the aim of the diffusion model is to learn to correlate different multi-modal conditions with their respective latent representation in the 3D GAN space.

The latent representation Next3D is $\mathbf{z}_i \in \mathbb{R}^{k \times d}$, which is a sequence of length $k$ consisting of embeddings of dimension $d$.
Therefore, we deviate from the 2D UNet \cite{unet} architecture that is used in image diffusion models adopted in \cite{gu2023learning}.
Instead, we build our model with 1D UNet $\epsilon_\theta$ where each residual block is followed by self-attention, linear, and cross-attention. 
We incorporate the multi-modal conditions into the denoising process through the cross-attention blocks. 
The conditional embeddings $\mathbf{c}_\mathcal{A}$ and $\mathbf{c}_\textit{v}$ are concatenated after applying dropout to produce a condition embedding $\mathbf{c} = [ \mathbf{c}_\textit{a}, \mathbf{c}_\textit{v}]$.
This condition embedding is cross-attended with the diffusion features as follows:

\begin{align}
    \text{{CrossAttention}}(Q_i, K_i, V_i) = \text{{Softmax}}\left(\dfrac{Q_i \ K_i^{T}}{\sqrt{{d_{k_i}}}}\right) \ V_i \enspace ,\\
    Q_i = f_i \ W_{Q_i}, \qquad K_i = \mathbf{c} \ W_{K_i}, \qquad V_i = \mathbf{c} \ W_{V_i}.
\end{align}
where $f_i$ are the diffusion features at block $i$, $d_{k_i}$ is the dimensionality of the keys, and $W$ are trainable projection matrices.

This architecture is considerably lighter than existing diffusion-based approaches.
In particular, our UNet has approximately $225$M parameters in comparison to the standard 2D diffusion model adopted by Control3Diff \cite{gu2023learning} with $595$M parameters, Stable-Diffusion~\cite{stable_diffusion} which has $1.45$B parameters or the even larger SDXL~\cite{podell2023sdxl} with $6.6$B parameters.

% All the parameters of $\epsilon_\theta, \mathbf{E}_\mathcal{I}$ and $\mathbf{E}_\mathcal{A}$ are trained end-to-end.

% One natural candidate for sequential data is using Transformers and initial exploration, we experimented with a DiT-inspired model simply concatenating the condition vector to the noise vector and denoising just the noise. 
% However, this model proved hard to train and required considerable GPU-compute. 

% As described in Section \ref{sec:dataset_generation}, our latents are . 
% Thus, we are forced to move away from the popular architectures, which rely on 2D-UNETs~\cite{unet}. 
% We choose to 

% We use a simple mechanism for conditioning which is borrowed from 2D-diffusion literature~\cite{Nichol2021GLIDETP, stable_diffusion} and we found effective in our case as well. 
% We flatten the 2D $\mathbf{c}_\mathcal{I}$ and concatenate it with $\mathbf{c}_\mathcal{A}$ along the sequence dimension. 
% Dropout as described in Section 3.2 is performed at this stage. 
% This condition sequence is used to generate the \textit{Key} and the \textit{Value} vectors for performing the Cross-Attention within the UNET. 
% The \textit{Query} vectors are generated from the latent features. We inject the condition at every step of the denoising process.

%%%%%%%%%%%%%%%%%%%%%%%%%%%%%%%%%%%%%%%%%%%%%%%%%%%%%%%%%%%%%%

\subsection{Training Pipeline}
Our approach requires training only the 1D diffusion model, unlike \cite{trevithick2023,gu2023learning,stable_diffusion}, which requires re-training the 3D GAN.
Instead, we use the pre-trained Next3D \cite{sun2023next3d} model, and we find that simple min-max normalization of the latents is sufficient.
We now explain how we generate the dataset to train our model and the implementation details.

%%%%%%%%%%%%%%%%%%%%%%%%%%%

\subsubsection{Dataset Generation}
\label{sec:dataset_generation}
%% Define the camera as well
% TODO from a camera $\pi_k$
For training our diffusion model, we require a dataset of the form $(\mathbf{x}_i, \mathcal{I}_i, \mathcal{M}_i, \mathcal{A}_i)$ where $\mathbf{x}_i$ is the latent of the 3D-GAN that generates the image $\mathcal{I}_i$, and whose segmentation map is $\mathcal{M}_i$ and attributes of the generated face are $\mathcal{A}_i$.
In order to generate the set of images $\mathcal{I}_i$, we need access to a camera distribution $\Pi$. 
We wish the camera distribution to reflect the distribution of the controlling images. 
Hence, we use the standard EG3D distribution with some noise to simulate different types of images we might encounter. 
In particular, we sample the FOV from a distribution which samples with uniform probability over $[22, 25]$ degrees with $70\%$ probability and $[18, 22]$ degrees with $30\%$ probability. 
Both the yaw and pitch are sampled from a uniform frontal facing distribution with a max deviation of $0.15$ radian, and the roll is set to zero. 
The camera radius is fixed at $2.7$ as we found the pre-trained checkpoint to be sensitive to the radius. 
We obtain the segmentation maps and attributes by running a face-segmentation network and the facial-attribute network from FaRL~\cite{zheng2022farl}. 
We generate a total of $6$M unique faces and their corresponding segmentation and facial attributes.

%%%%%%%%%%%%%%%%%%%%%%%%%%%

\subsubsection{Masking Scheme} 
\label{sec:masking}
In order for our model to restrict the edits to specific regions, we perform masking of all the modalities. 
For the attributes, we simply replace each of the attributes with a special value $\texttt{MASK}_\textit{a}$. 
For the visual attributes, we independently mask the segmentation map and the RGB condition with a brush-like mask filling in the masked region with value $\texttt{MASK}_\textit{v}$. 
Furthermore, we mask the RGB and the segmentation map by dropping regions corresponding to a randomly chosen class. 
The independent probability of masking any of the modalities is 90\%.

%%%%%%%%%%%%%%%%%%%%%%%%%%%
\subsubsection{Implementation Details}
% We found out that the model is rather robust with respect to the training hyperparameters. 
All components $\epsilon_\theta, \mathbf{E}_\textit{a}$ and $\mathbf{E}_\textit{v}$ are trained end-to-end using the loss in Eq (\ref{eq:l_diff2}).
The diffusion model can be trained with both $\mathbf{x}_0$-prediction and $\mathbf{v}$-prediction paradigms, but for the results presented in this paper, we employed the $\mathbf{v}$-prediction approach with $1000$ denoising steps.
We use a Cosine scheduler \cite{nichol2021improved} with $\beta_{start}=0.008$ and $\beta_{end}=0.02$
The model is trained for $500k$ steps with a batch size of 512 over 4 V100 GPUs or a single A100 GPU. 
We used the ADAM optimizer with the OneCycle LR scheduler with no weight decay, a maximum learning rate of $1\times10^{-4}$, and warm-up for the first $10k$ step.
To be able to employ Classifier-Free Guidance during sampling, we drop each of the conditions $20\%$ of the time during training.

%%%%%%%%%%%%%%%%%%%%%%%%%%%%%%%%%%%%%%%%%%%%%%%%%%%%%%%%%%%%%%

\subsection{Sampling}
\label{sec:sampling}
We use two forms of conditioning, visual and attribute-based, so our modified noise estimate for Classifier-Free Guidance (CFG) sampling in Eq (\ref{eq:cfg}) becomes:
\begin{equation}
\begin{split}
    \tilde{\epsilon}_\theta(\mathbf{z}_t, \mathbf{c}_\textit{v}, \mathbf{c}_\textit{a}) &= {\epsilon}_\theta(\mathbf{z}_t, \phi, \phi) \\
     &+ \omega_\textit{v}[ {\epsilon}_\theta(\mathbf{z}_t, \mathbf{c}_\textit{v}, \phi) - {\epsilon}_\theta(\mathbf{z}_t, \phi, \phi)] \\
     &+ \omega_\textit{a}[ {\epsilon}_\theta(\mathbf{z}_t, \mathbf{c}_\textit{v}, \mathbf{c}_\textit{a}) - {\epsilon}_\theta(\mathbf{z}_t, \mathbf{c}_\textit{v}, \phi)]
\end{split}
\label{eq:sampling}
\end{equation}
where $\omega_\textit{v}$ and $\omega_\textit{a}$ are the visual and the attribute weights, respectively, and we drop timestep $t$ for simplicity.

We use standard DDIM sampling~\cite{song2021denoising} with $T=100$ steps as it provides a good balance between sample quality and speed.
When performing edits, it is critical to maintain a trade-off between reconstruction and edits. 
Therefore, we found it useful to break the editing process into two stages: 1. Reconstruction stage and 2. Editing stage. 
In the reconstruction stage, we do not mask any of the attributes as explained in Section \ref{sec:masking}, and we perform the denoising process in Eq (\ref{eq:sampling}) until some timestep $t_{rec} < T$. 
In the editing stage, we mask out the modality to be edited and continue sampling from $t_{rec}+1$ to $T$. 
This design choice guarantees that different conditioning modalities do not lead $\epsilon_{\theta}$ into different contradictory directions leading to unusable sampling. 

The whole sampling process takes $0.25$ seconds per generated latent, and we are able to generate a batch size of 32 latents in approximately $8$ seconds on a single A100 GPU.
This generation time does not include the time taken to encode the conditions, as they are pre-computed once throughout the denoising process and introduce only a small overhead.

%%%%%%%%%%%%%%%%%%%%%%%%%%%%%%%%%%%%%%%%%%%%%%%%%%%%%%%%%%%%%%

\begin{figure}[!t]
    \centering
     \includegraphics[width=\linewidth]{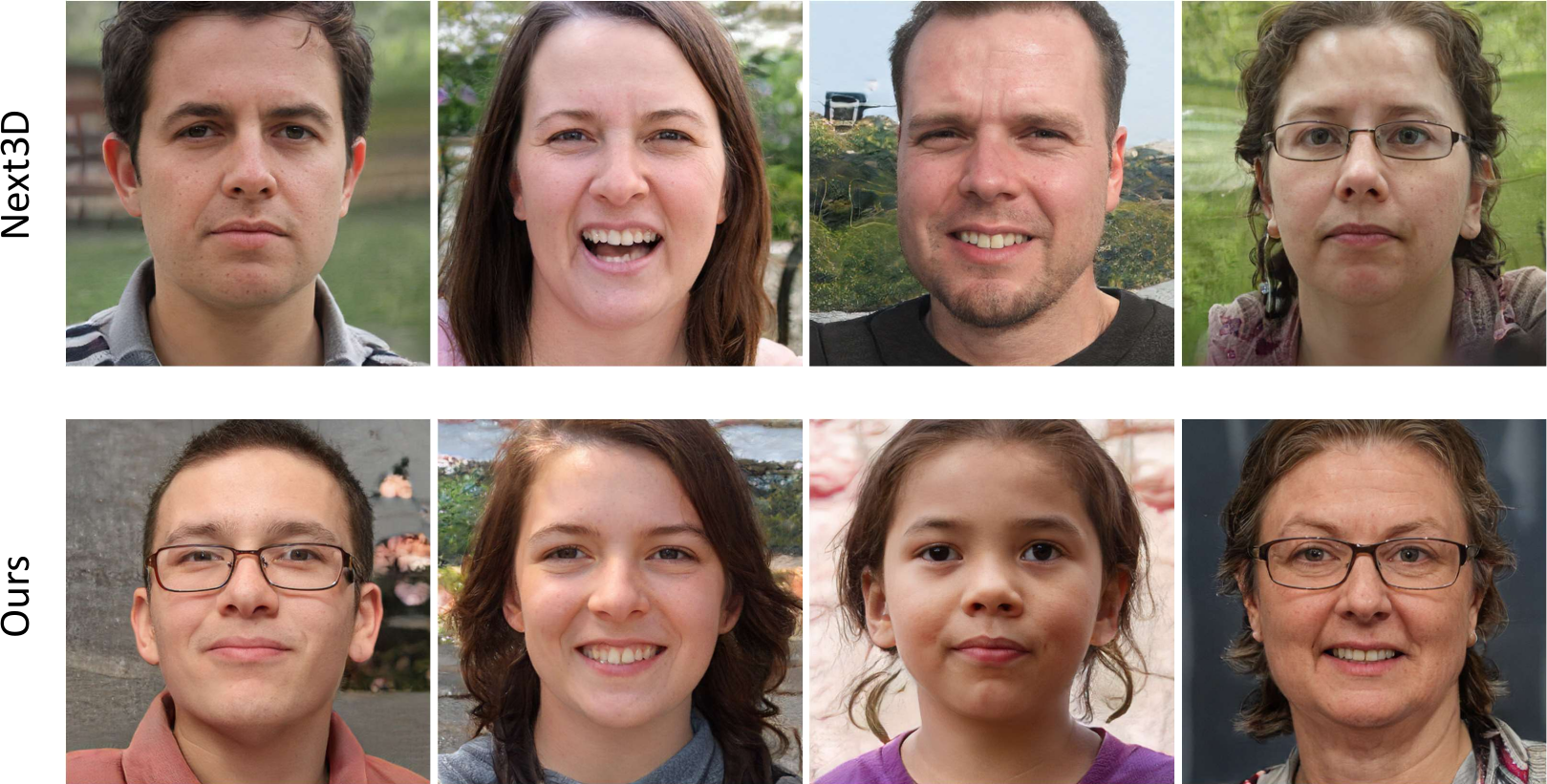}
    \caption{ \captionsize \textbf{Unconditional generation.} Our approach maintains the performance of the base GAN, Next3D \cite{sun2023next3d}, and can generate high-quality avatars in the unconditional setting.}
    \label{fig:uncond}
    \vspace{-0.5cm}
\end{figure}

\section{Experiments and Results}
\label{sec:experiments}

To demonstrate the capabilities of our proposed approach, we evaluate it on the tasks of multi-conditional avatar generation and editing.
To the best of our knowledge, there exists no other comparable approach with the same capabilities.
Therefore, we design an optimization-based baseline inspired by \cite{gu2023learning,nie2021controllable} that we denote as $\mathcal{W}^{++}$, and we compare against it.
We provide qualitative comparisons on multi-conditional avatar generation and editing. 
For the quantitative comparison, it is infeasible to cover all combinations of possible conditions.
Therefore, we choose two tasks where we compare against the baseline in terms of some quality metrics.

% Note that we can not generate results with multi-modal conditions using the baseline that we described above due to \TODO{}.
% However, we provide this comparison for the editing task.

%%%%%%%%%%%%%%%%%%%%%%%%%%%%%%%%%%%%%%%%%%%%%%%%%%%%%%%%%%%%%%

\subsection{Baseline}
The baseline is purely GAN-based, and it follows the standard image inversion pipeline ~\cite{Abdal2019Image2StyleGANHT,sg++}, where only reconstruction losses $\mathcal{L}_{rec}$ are used. Specifically, given input image $\mathcal{I}$ and a pre-trained GAN generator \textsf{G}, we try to find a latent $w^+$ such that: 
\begin{equation}    
w^+ = \arg \min_x \mathcal{L}_{rec}(\text{\textsf{G}}(x)\enspace, \mathcal{I}) 
\end{equation}
We adapt this objective to incorporate multiple conditions as follows: 
\begin{multline}
    w^+ = \arg \min_x [ \mathcal{L}_{rec}(\text{\textsf{G}}(x), \mathcal{I}) + \lambda_{attr}\mathcal{L}_{attr}(\mathbf{A}(\text{\textsf{G}}(x)),  \mathcal{A}_{target})\\
     + \lambda_{seg}\mathcal{L}_{seg}(\mathbf{S}(\text{\textsf{G}}(x)),  \mathcal{S}_{target})],
\end{multline}
where $\mathbf{A}$ and $\mathbf{S}$ are the differentiable attribute and segmentation prediction networks, respectively from FaRL \cite{zheng2022farl}. 
The losses $\mathcal{L}_{attr}$ and $\mathcal{L}_{seg}$ perform attribute-based edits and segmentation-based edits, while $\mathcal{L}_{rec}$ controls the RGB reconstruction quality. 
The L2-norm is used for all these losses.

% This baseline is very strong, as the optimization procedure checks the error with respect to the target at every step, while our method only encodes the target once and the sampling process does not take the rendering process of the GAN into account.

% However, we can still validate the performance of our model by measuring its generative performance. 
% Additionally, we can measure the \textit{alignment} of the generated samples with respect to the input conditions.
\begin{figure}[!t]
    \centering
     \includegraphics[width=\linewidth,]{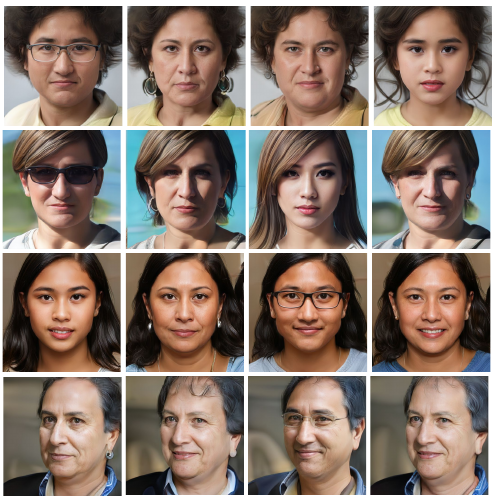}
    \caption{ \textbf{Multiple completions.} As sampling from a diffusion model is inherently probabilistic, we can generate multiple samples adhering to the same conditions. In each row, we fix the hair above the forehead and generate multiple samples corresponding to that hair patch. These images are generated with the same starting noise and the DDIM parameter $\eta$ set to $0.25$.}
    \label{fig:diversity}
\end{figure}

%%%%%%%%%%%%%%%%%%%%%%%%%%%%%%%%%%%%%%%%%%%%%%%%%%%%%%%%%%%%%%

\subsection{Avatar Generation Results}
\begin{figure}[!t]
    \centering
     \includegraphics[width=\linewidth]{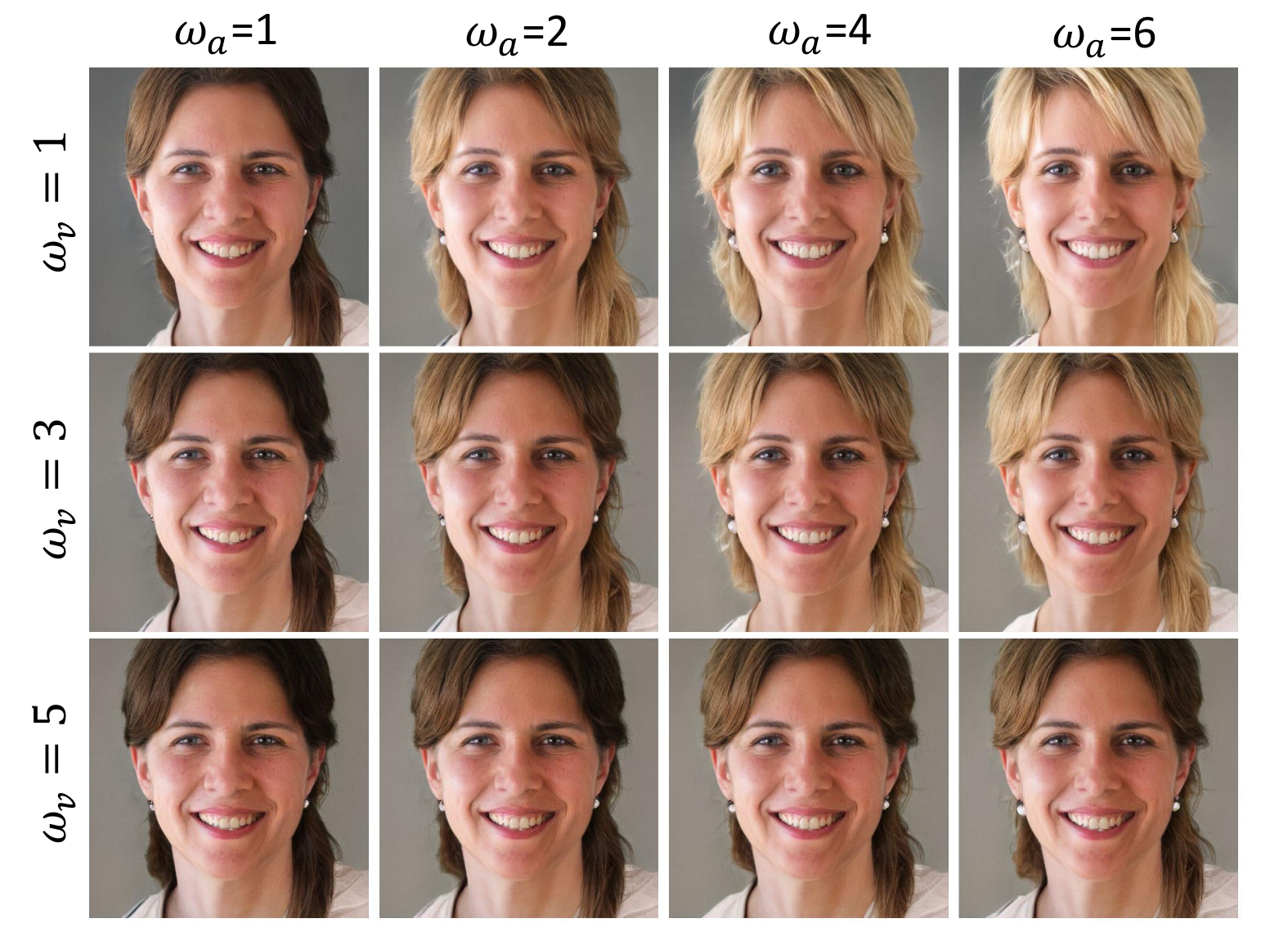}
    \caption{ \captionsize \textbf{Controlled generation.}  We can control the relative importance of each of the modalities by varying the corresponding weights, $\omega_{a}$ for attributes, and $\omega_{v}$ for the visual conditions. Greater weights mean the samples retain larger adherence to the corresponding condition. Our model is able to generate in-domain samples within a large range of numerical values of the weights.}
    \label{fig:weights}
\end{figure}

\noindent \textbf{Unconditional Generation}
Our approach employs the pre-trained 3D GAN, Next3D \cite{sun2023next3d} as a generator that is kept frozen during training as illustrated in \Cref{fig:pipeline}.
It is essential to ensure that the quality of the generated avatars by our model remains largely consistent with those produced by the \emph{unconditional} base GAN, Next3D.
To examine this, we set all conditions to our model to the \emph{null-condition} to perform unconditional generation.
\Cref{fig:uncond} shows some examples of samples generated with no condition.
We note that the samples generated are diverse and retain the high-quality generation of the underlying 3D GAN.
To validate this further, we generate 50k unconditional samples using our approach, and we compute the FID score \cite{fid_nips2017} between the generated samples and the training set generated by Next3D as described in \Cref{sec:dataset_generation}.
We achieved an FID score~\cite{fid_nips2017} of 7.43, which suggests that the generated samples from our approach follow a similar distribution of samples from the base GAN.

%%%%%%%%%%%%%%%%%%%%%%%%%%%

\noindent \textbf{Diversity}
Diffusion models are capable of generating diverse samples by following different denoising trajectories \cite{dhariwal2021diffusion}.
Ideally, we would like our model to inherit this capability and generate diverse samples for a given condition.
\Cref{fig:diversity} shows some examples where we condition our model on the RGB image of the hair.
The figure demonstrates that our model can generate diverse samples that adhere to the given condition.

%%%%%%%%%%%%%%%%%%%%%%%%%%%

\begin{figure*}[!t]
    \centering
     \includegraphics[width=0.95\linewidth]{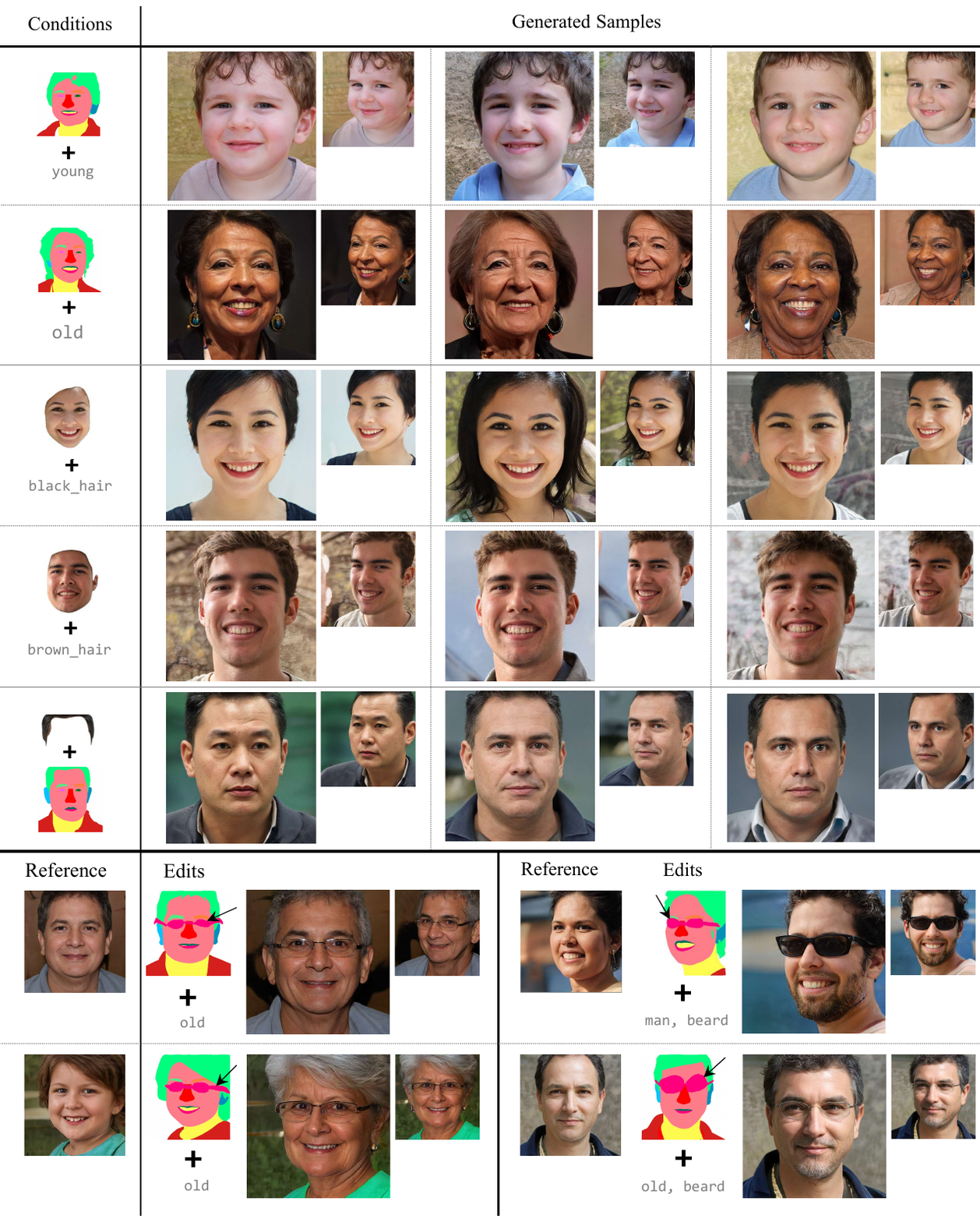}
    \captionof{figure}{\textbf{Multi-modal conditional generation and editing.} Our approach can perform generation and editing, giving any combination of the multi-modal conditions. [Black arrows point to edited parts in the segmentation map]}
    \label{fig:qual_1}
\end{figure*}

\noindent \textbf{Multi-Conditional Generation}
We showcase the multi-modal conditioning capabilities of our method in \Cref{fig:qual_1}.
Given a face segmentation map and a specific attribute as conditions, our method can generate high-quality and diverse samples that adhere to the conditions.
For the unspecified attributes, the model is free to generate samples with any arbitrary attributes.
For example, when an RGB image of the face and the hair color attribute are provided, our model preserves the face and generates different styles of hair that have the specified color. The important characteristic of our approach is that the same model works without conditioning information, all types of conditioning information, or arbitrary subsets of types of conditioning information.

\begin{figure}[!t]
    \centering
     \includegraphics[width=\linewidth]{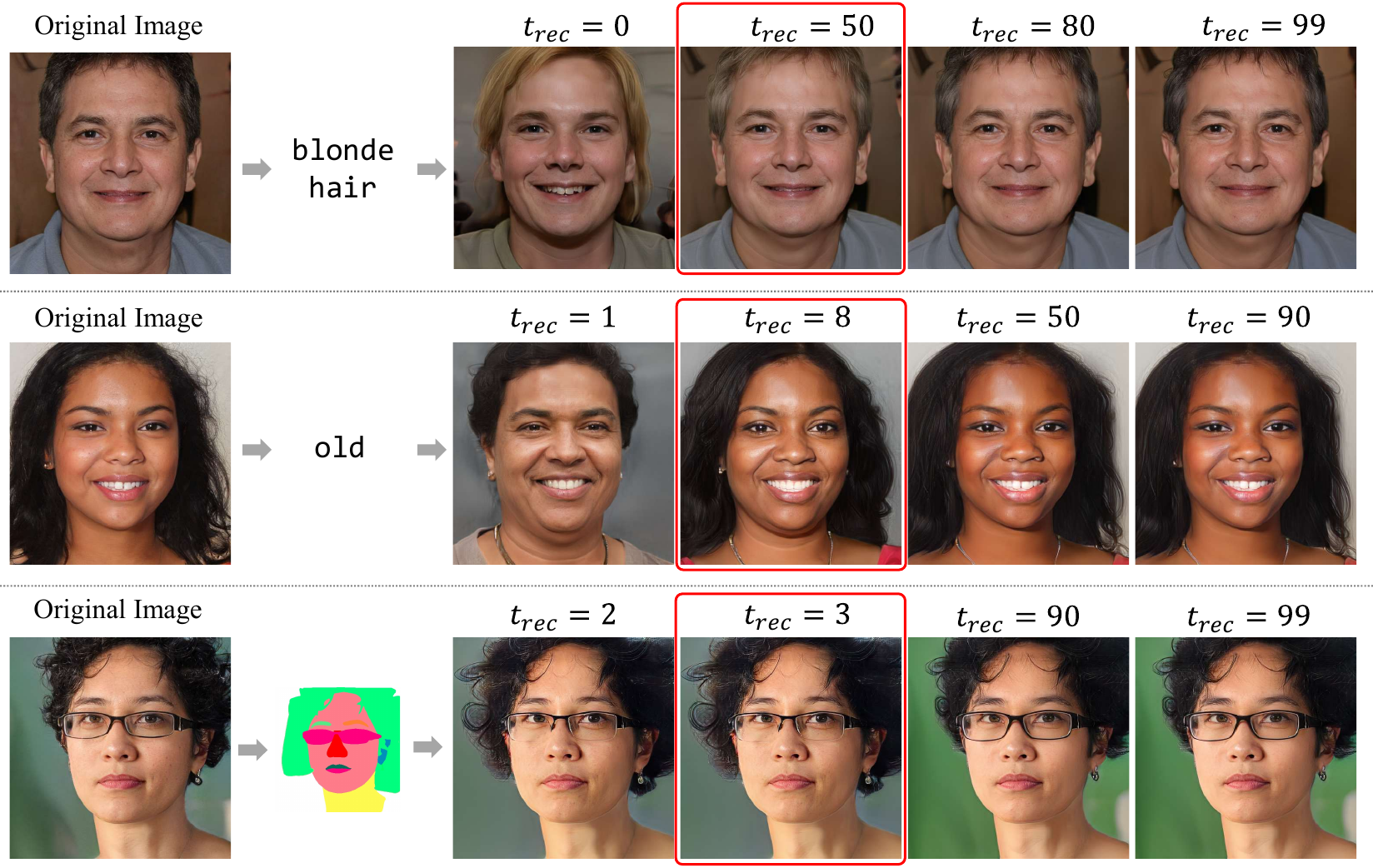}
    \captionof{figure}{\textbf{Illustration of the effect of choice of $t_{rec}$.} Our editing strategy requires us to choose a timestep $t_{rec}$ depending on the applied edit. Applying the edit too early causes the generation to vary significantly from the input image while applying the edit too late makes the model disregard the condition. The best result is marked with a red box.}
    \label{fig:t_rec}
\end{figure}

%%%%%%%%%%%%%%%%%%%%%%%%%%%%%%%%%%%%%%%%%%%%%%%%%%%%%%%%%%%%%%

\begin{figure*}[!t]
    \centering
     \includegraphics[width=\linewidth, height=4cm]{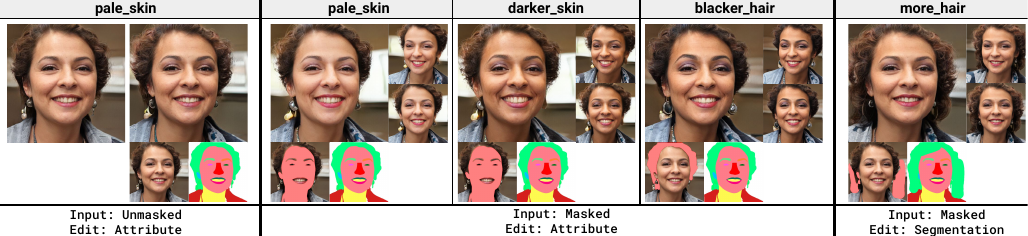}
    \caption{\textbf{Masking for editing.} During inference, the proper modality should be masked in order to perform the edit. (\textbf{Left}) When the hair in the RGB is left unmasked and the edit for \texttt{pale\_skin} is performed, the generated image does not follow the requested edit. (\textbf{Others}) With proper masking, the edit is transferred to the generated image for both the hair and the skin. Additionally, in the last column, note that in order for the edit on the segmentation to propagate to the generated sample, the corresponding RGB is masked. For each example, we show the masked part of the RGB in {pink}.}
    \label{fig:masking}
\end{figure*}

%%%%%%%%%%%%%%%%%%%%%%%%%%%%%%%%%%%%%%%%%%%%%%%%%%%%%%%%%%%%%%

\subsection{Avatar Editing Results}
\label{sec:exp_edit}

Traditionally, GAN-based editing relies on first finding a latent code that can generate the input image and then moving in the latent space in certain global or local directions to apply the edit.
We try to replicate this process using the idea of the \emph{Reconstruction} stage and the \emph{Editing} stage as described in \Cref{sec:sampling}.
We provide a comparison against the baseline $\mathcal{W}^{++}$ described above.

% Our framework can perform these edits individually - using only the attributes, segmentation map or the RGB, or jointly using any combination thereof. 
% The key idea is that we mask out the the modality which is influenced by the edit. An illustration of this is shown in Fig.~\ref{fig:masking}.

%%%%%%%%%%%%%%%%%%%%%%%%%%%

\begin{figure*}[!t]
    \centering
     \includegraphics[width=0.95\linewidth]{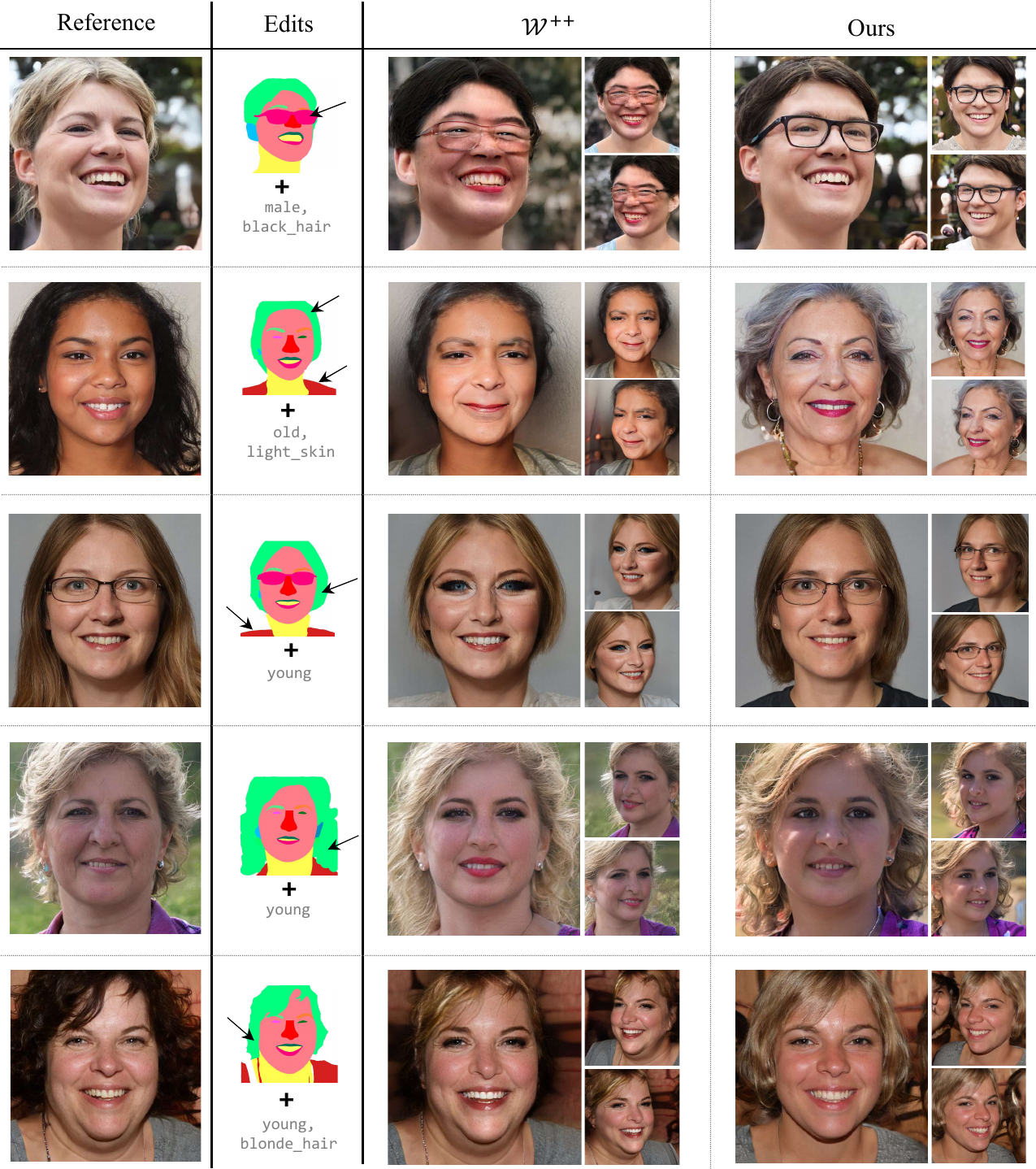}
    \captionof{figure}{\textbf{Comparison against the baseline.} Our approach can perform edits given multi-modal conditions, while the baseline $\mathcal{W}^{++}$ struggles to produce realistic edits. Note specifically that the baseline particularly has troubles with \texttt{glasses}, generating implausible \texttt{glasses} (Row 1), or failing to preserve them (Row 3). Our model has approximate segmentation adherence, unlike the baseline, which is more accurate. This is because $\mathcal{W}^{++}$ involves the rendering process while we work in the latent space. However, this adherence comes at the cost of realism (Rows 2, 3, 5), while our method generates samples of significantly higher quality. [Black arrows point to edited parts in the segmentation map]}
    \label{fig:qual_2}
\end{figure*}

\hfill \break
\noindent \textbf{Qualitative Comparison }
\Cref{fig:qual_2} provides a qualitative comparison against the baseline on multi-conditional avatar editing.
Generally, our approach performs remarkably well on attribute editing, such as gender, age, skin, and hair color. 
It also succeeds in applying segmentation mask edits, including adding glasses, editing hairstyles, and adding clothing. %\todo{shoulders}.
On the other hand, the baseline struggles to produce realistic samples that fulfill the edit or to apply the edits in some cases.
Moreover, the generated samples have several artifacts especially when glasses are present in the sample. 
However, it manages to follow the segmentation map to a good extent.

% While the baseline is able to actually perform multimodal edits, the quality of the edits is subpar. 
% Consider changing the color of the hair - the baseline has most of the information coming from the RGB itself. 
% Consequently, the reconstruction is quite good, but the edits are only applied partially. 
% Increasing $\lambda$ for the edit can lead to images with significant artifacts. 
% Our method is able to retain the details of the hair, as the Reconstruction stage itself encodes the structure in the latent. 
% The optimization-based method is unable to retain such details during the course of the optimization. 
% The quantitative results reinforce this - in all metrics, apart from actual reconstruction, our method outperforms the baseline, especially in realism. 
% The inversion methods are designed for reconstruction and hence outperform our method, but our method requires no backpropagation and hence can be used on cheap commodity hardware, while optimization-based methods require more computation.

%%%%%%%%%%%%%%%%%%%%%%%%%%%

\hfill \break
\noindent \textbf{Quantitative Comparison}
We provide a quantitative comparison for two settings. The first is conditional generation given an RGB image of only the face, a hair segmentation map, and a hair color attribute.
For the second setting, we provide half of the RGB image, the other half of the segmentation map, and a null attribute.
We generate 1k samples for both experiments.
The results for these two experiments are reported in \Cref{tab:quant}.
For evaluating the reconstruction quality, we report the PSNR, (Multi-scale) SSIM, and LPIPS~\cite{Zhang_2018_CVPR}. 
To verify that the edit has been applied without significantly changing the identity, we report the cosine distance between the Arcface embeddings ~\cite{arcface} of the edited and the input image (ID).
Acc measures the strength of the edit and is the L1 difference between the attribute set during sampling, and the attribute predicted from the generated image. 
mIOU performs the role of measuring the adherence to the provided segmentation map, and it is the mean intersection-over-union between the segmentation labels provided during sampling and the labels predicted by the segmentation network on the generated image. 
For measuring both Acc and mIOU, we use FaRL \cite{zheng2022farl}.

For the first experiments, our approach outperforms the baseline in terms of image quality and preserving the identity.
On the other hand, the baseline $\mathcal{W}^{++}$ has better adherence to the segmentation map and the global attributes as it is optimized over the predictions from the segmentation and the attribute network.
However, this comes at a cost of significant changes in the face identity, as measured by ID (See \Cref{fig:qual_2}). 
For the second experiment, our model performs on par with the baseline despite being orders of magnitude faster than the baseline.

\begin{table}
    \centering
    \begin{tabular}{|c|c|c|c|c|c|c|}        
        \hline
        \multicolumn{7}{|l|}{\textbf{Conditions:} \quad RGB: Face, \quad SEG: Hair, \quad Attr.: Hair Color} \\
        \hline
        & PSNR $\uparrow$ & SSIM $\uparrow$ &  LPIPS $\downarrow$  & ID $\uparrow$ & mIOU $\uparrow$ & Acc $\uparrow$ \\
        \hline
        $\mathcal{W}^{++}$ & 19.38  & 0.80 & 0.14 & 0.48 &  \textbf{82.41} & \textbf{0.05} \\
        \textbf{Ours} & \textbf{22.18} & \textbf{0.89} & \textbf{0.08} & \textbf{0.70} & 79.62 & 0.35 \\     
        \hline
        \multicolumn{7}{|l|}{\textbf{Conditions:} \quad RGB: Left half, \quad SEG: Right half, \quad Attr.: Null} \\
        \hline
        & PSNR $\uparrow$ & SSIM $\uparrow$ & LPIPS & ID $\uparrow$ & \multicolumn{2}{c|}{Time $\downarrow$} \\
        \hline
        $\mathcal{W}^{++}$ & \textbf{24.16} & \textbf{0.93} & 0.07 & \textbf{0.81} & \multicolumn{2}{c|}{90s} \\
        \textbf{Ours} & 23.97 & 0.91 & 0.07 & 0.79 &  \multicolumn{2}{c|}{\textbf{0.25s}} \\     
        \hline
    \end{tabular}
    \caption{A quantitative comparison between our proposed approach and the baseline on multi-conditional avatar generation. PSNR, SSIM, and LPIPS are computed only for the RGB provided as condition. Our model is able to perform edits without modifying the identity while being much faster. Time is measured on a single A100 GPU.}
    \label{tab:quant}
    % \vspace{}
\end{table}

%%%%%%%%%%%%%%%%%%%%%%%%%%%%%%%%%%%%%%%%%%%%%%%%%%%%%%%%%%%%%%

\subsection{Ablation Study }
\noindent \textbf{Sampling Weights}
We examine how the weights $\omega_\textit{v}$ and $\omega_\textit{a}$ in \Cref{eq:sampling} affect the generated results samples in \Cref{fig:weights}.
In this example, the full face is provided as an RGB condition as well as ``\texttt{blonde\_hair}'' attribute.
When setting $\omega_\textit{v}$ high, the face is preserved and prioritized over the attribute.
On the contrary, a higher $\omega_\textit{a}$ prioritizes the attribute over the identity of the face.
Therefore, these hyperparameters can be used for more control over the generation.

%%%%%%%%%%%%%%%%%%%%%%%%%%%

\hfill \break
\noindent \textbf{Choice of $t_{rec}$}
When performing edits, a key hyperparameter is $t_{rec}$, which corresponds to the number of denoising steps taken under the groundtruth conditions rather than the editing ones. 
With $t_{rec} = 0$, the reference image does not affect the sampling process, and hence we are basically sampling under the editing conditions. 
On the other hand, when $t_{rec} = T$, the editing conditions do not affect the sampling process.
\Cref{fig:t_rec} shows editing examples under different choices of $t_{rec}$.
We make the following observations. 
First, for local attributes, \eg hair color, a small $t_{rec}$ would change the identity of the person, while a higher one would barely apply the edits. 
An intermediate $t_{rec}$ achieves a good trade-off in this case.
Second, for global attributes that change the entire face, \eg young, and old, it is enough to do reconstruction for some steps, \eg $10-20 \%$ of the total steps, to preserve the identity.
Finally, for edits that are based on a segmentation map, for instance, making the hair longer, also few steps are sufficient to align the sampling trajectory to the input image.
We find this to be intuitive since, in case of editing the hair color, fewer editing steps are needed to change only the color, while for global changes such as the age, more editing steps are needed to change the structure of the avatar.
These observations are consistent with several diffusion-based image editing approaches \cite{hertz2022prompt,cao2023masactrl}.

%%%%%%%%%%%%%%%%%%%%%%%%%%%

\hfill \break
\noindent \textbf{Masking for Editing}
In Figure~\ref{fig:masking}, we provide an illustration of how we perform masking to make sure that the generated samples correspond to the desired edit.
The figure shows that masking is essential to achieve the desired edits.
When applying attribute editing, the affected regions need to be masked out. For instance, when changing the skin color, the whole face needs to be masked.
This mechanism guides the diffusion process to focus on the relevant regions, similar to image inpainting.

%%%%%%%%%%%%%%%%%%%%%%%%%%%

%%%%%%%%%%%%%%%%%%%%%%%%%%%%%%%%%%%%%%%%%%%%%%%%%%%%%%%%%%%%%%

\section{Conclusion and Future Work}
\label{sec:conclusion}
We proposed an approach for 3D avatar generation and editing with multi-modal conditioning.
We employed a latent diffusion model to impose control over the latent space of a pre-trained 3D GAN.
The proposed model is lightweight and fast to train and sample from, as it does not require re-training the GAN.
Experiments demonstrated that our approach can produce high-quality avatars that adhere to any set of given conditions.
Moreover, it can perform high-fidelity editing while preserving the identity of the reference avatar compared to a solely GAN-based baseline.
A limitation of our work is that we inherit biases from the training data and training methods used to generate the initial head avatar model.

For future work, there are many possible extensions to this work.
First, the model can be extended to support more conditioning modalities such as sketches and landmarks. Second, we would like to build a joint model for controlling animation and head avatars via diffusion.

% Second, the hyperparameters $\omega_v$ and $\omega_a$

% One line of inquiry is how to improve the reconstruction quality. 
% If we had access to the underlying mesh and the correct camera pose, we could use guided diffusion using a simple photometric loss. 
% In theory, our model can predict both the mesh parameters and the camera parameters in the proposed framework simply by appending those parameters to the sequence and using them as denoising targets. 
% An orthogonal research direction is on improving generalization. 
% The proposed framework work remarkably well when generating samples and using multi-modal data from GAN generated images. 
% However, in order to be useful in the real-world, we would require inversion of real images. We did not explore that direction in the current work. 
% Furthermore, our model would inherit the biases of both the original data the underyling GAN was trained on and the synthetically generated data. 

% Going past these limitations however, our work opens up a large research field - given that we show how to efficiently sample in the space of 3D-GANs, what is the most effective latent space, which latent space is the most disentangled, and how to design better sampling algorithms that incorporate human feedback.

% \FloatBarrier
\newpage
{
    \small
    \bibliographystyle{ACM-Reference-Format}
    \bibliography{main}
}
% WARNING: do not forget to delete the supplementary pages from your submission 

\clearpage
% \setcounter{page}{1}
% \maketitlesupplementary

\appendix
\twocolumn[\Huge\sffamily Supplementary Material \\ AvatarMMC: 3D Head Avatar Generation and Editing with Multi-Modal Conditioning]

\section{Additional Qualitative Results}
\begin{figure*}[thb]
    \centering
     \includegraphics[width=0.95\linewidth]{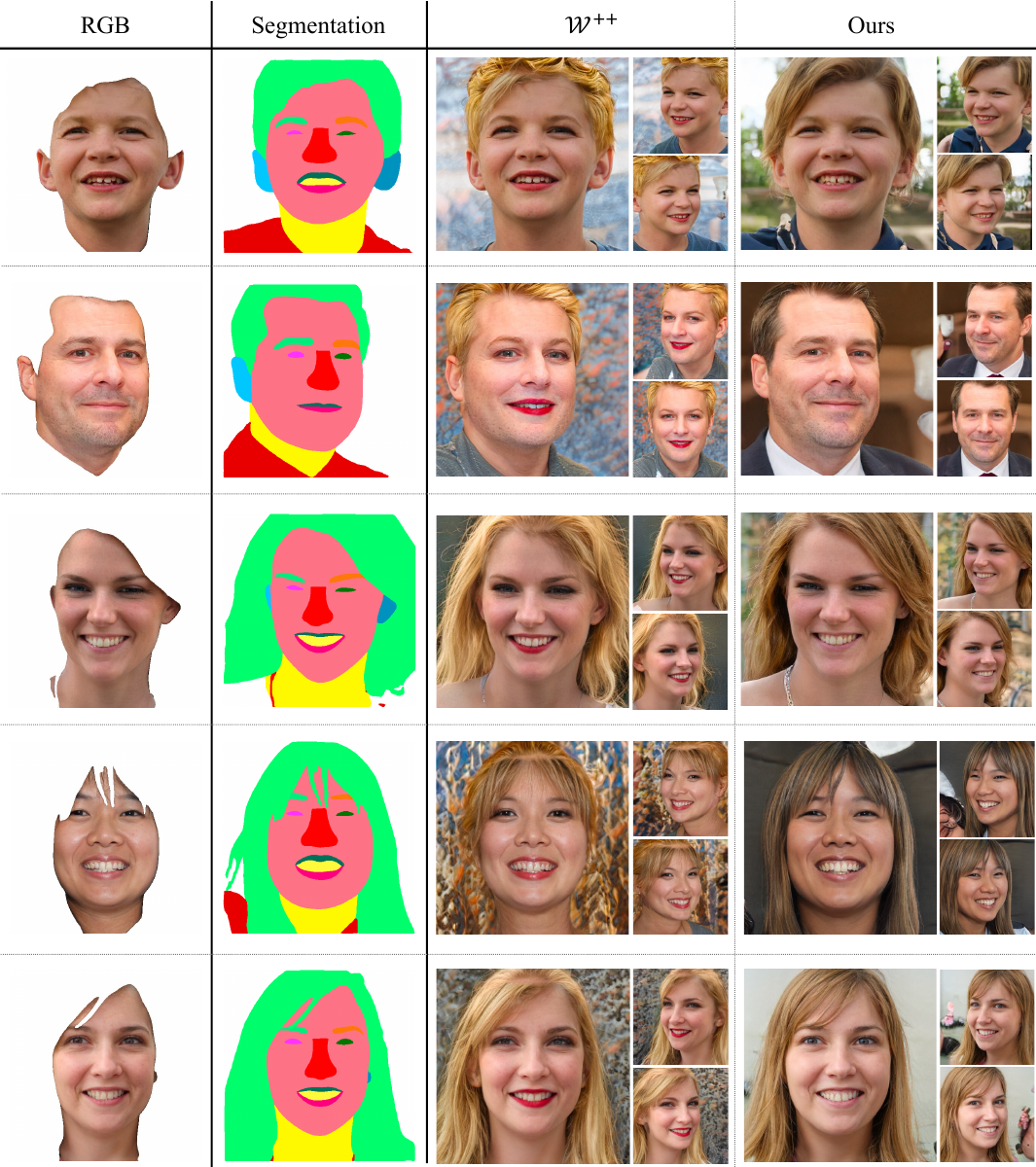}
    \captionof{figure}{\textbf{Qualitative comparison to the baseline on multi-modal generation.} Given the face-region RGB, the hair-segmap, and attribute blonde, our method generates avatars that are more natural than the baseline. The baseline applies the requested modality more aggressively than the proposed method (See cf. Tab. \ref{tab:quant} last column). However, the realism, and identity preservation of the generated avatars is lower when compared to the proposed method.}
    \label{fig:qual_1_suppl}
\end{figure*}

\begin{figure*}[thb]
    \centering
     \includegraphics[width=0.95\linewidth]{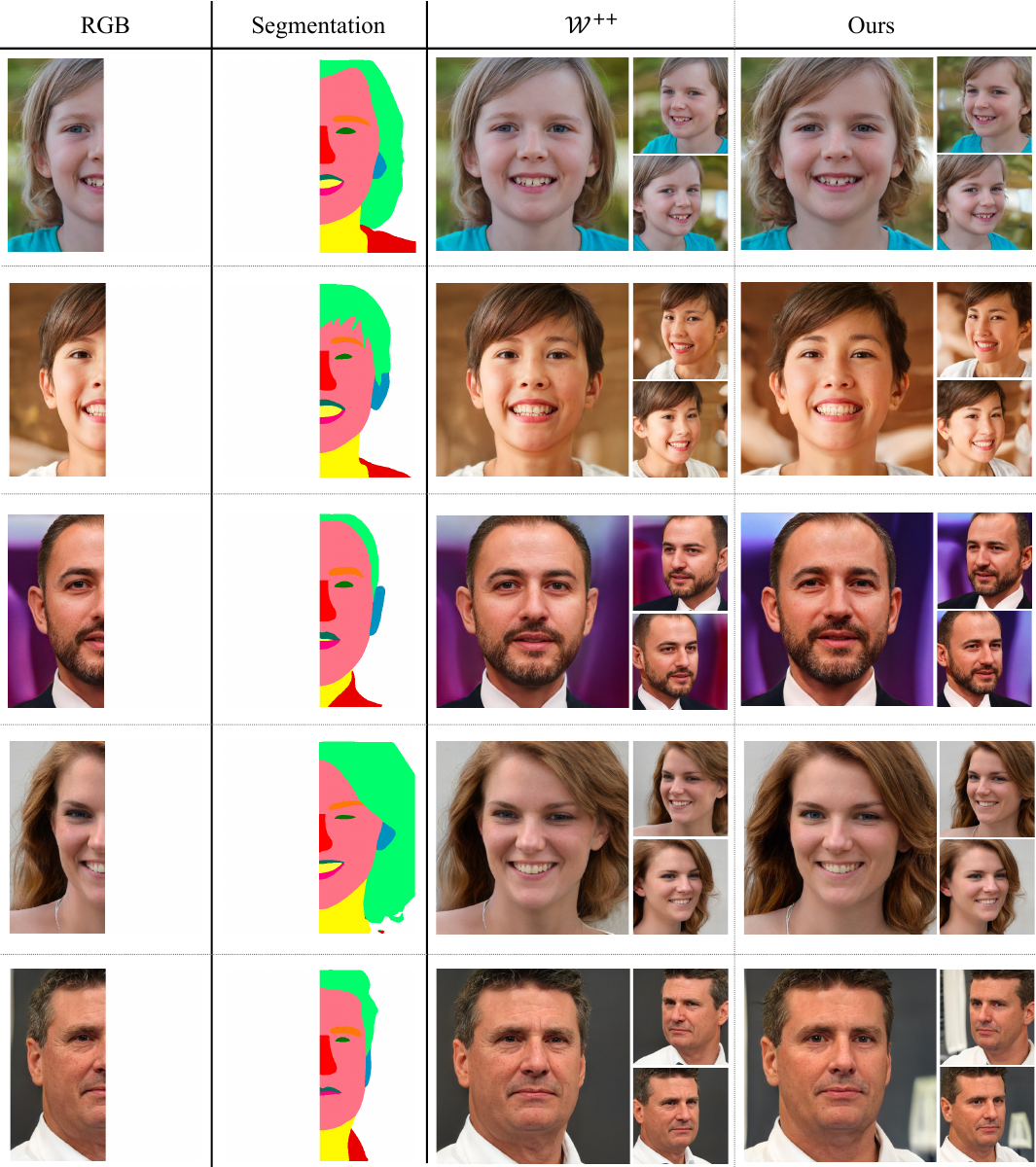}
    \captionof{figure}{\textbf{Qualitative comparison to the baseline.} Our approach can perform reconstruction as well as the baseline (See cf. Tab. \ref{tab:quant}), while being substantially faster. The results are multi-view consistent. Even without access to the renderer, our method is able to faithfully reconstruct fine details, such as the moustache in the last row.}
    \label{fig:qual_2_suppl}
\end{figure*}

We provide some qualitative examples in \Cref{fig:qual_1_suppl,fig:qual_2_suppl} for the two experiments described in \Cref{sec:exp_edit} and reported in \Cref{tab:quant}.
\Cref{fig:qual_1_suppl} shows that the optimization-based baseline produces avatars with lower realism, and has a tendency to correlate blondness with red lips.
On the other hand, our approach produces more realistic results and preserves the identity better than the baseline.
\Cref{fig:qual_2_suppl} shows that for the second experiment, our approach performs comparably well with the baseline while being orders of magnitude faster.

%%%%%%%%%%%%%%%%%%%%%%%%%%%%%%%%%%%%%%%%%%%%

\section{Detailed Architecture of the 1D Diffusion UNet}
\label{sec:appendix_arch}

We organize our UNET in the same manner as SD~\cite{stable_diffusion}. 
Concretely, the diffusion timestep is projected by a \texttt{SinCos-Lin} \texttt{ear-GELU-Linear} block to the same dimension as the UNET to create a timestep embedding.
We calculate a per-channel shift $\gamma$ and scale $\beta$ from the timestep embedding to modulate the UNET activations $\mathbf{h}$ as $\mathbf{h} \leftarrow [\mathbf{h} * (1 + \beta)] + \gamma$.

At each layer in the UNet, we have two residual blocks followed by self and cross-attention blocks. 
Each residual block in turn is composed of two \texttt{Conv1D-GN-SiLU} units, where the first block performs the scaling and shifting of the timestep embedding. The residual path has a \texttt{Conv1D} operation.

The base number of channels is $512$. We use three UNet downsampling blocks with a channel multiplier of $1,2,4$.
Further, we use skip connections in the UNet between the downsampling and upsampling blocks.

% \begin{table*}[h!]
% \centering
% \begin{tabular}{c|c|c|c|}
% \cmidrule[0.8pt]{2-4}
%  & \multicolumn{1}{|c|}{Representation} & \multicolumn{1}{|c|}{Animatable} & \multicolumn{1}{|c|}{Prior}\\ \cline{1-4} 
%  \multicolumn{1}{|c|}{Ours} &  \multicolumn{1}{|c|}{Ours} &  \multicolumn{1}{|c|}{Ours} &  \multicolumn{1}{|c|}{Ours}\\ \midrule
%   \multicolumn{1}{|c|}{Ours}&  \multicolumn{1}{|c|}{Ours}&   \multicolumn{1}{|c|}{Ours}&  \multicolumn{1}{|c|}{Ours} \\ \midrule
%   \multicolumn{1}{|c|}{Ours}&  \multicolumn{1}{|c|}{Ours} &  \multicolumn{1}{|c|}{Ours} &  \multicolumn{1}{|c|}{Ours} \\ \midrule
%  \multicolumn{1}{|c|}{Ours} &  \multicolumn{1}{|c|}{Ours} & \multicolumn{1}{|c|}{Ours}  &   \multicolumn{1}{|c|}{Ours}\\ \bottomrule
% \end{tabular}
% \caption{Your caption here}
% \label{tab:my_label}
% \end{table*}

\end{document}